# A Corpus-based Evaluation of a Domain-specific Text to Knowledge Mapping Prototype

Rushdi Shams
Department of Computer Science and Engineering, Khulna University of Engineering & Technology (KUET), Bangladesh
Email: rushdecoder@yahoo.com

Adel Elsayed
M3C Research Lab, the University of Bolton, United Kingdom
Email: a.elsayed@bolton.ac.uk

Quazi Mah- Zereen Akter
Department of Computer Science & Engineering, University of Dhaka, Bangladesh
Email: mahzereen@yahoo.com

*Abstract*— **The aim of this paper is to evaluate a Text to Knowledge Mapping (TKM) Prototype. The prototype is domain-specific, the purpose of which is to map instructional text onto a knowledge domain. The context of the knowledge domain is DC electrical circuit. During development, the prototype has been tested with a limited data set from the domain. The prototype reached a stage where it needs to be evaluated with a representative linguistic data set called corpus. A corpus is a collection of text drawn from typical sources which can be used as a test data set to evaluate NLP systems. As there is no available corpus for the domain, we developed and annotated a representative corpus. The evaluation of the prototype considers two of its major components- lexical components and knowledge model. Evaluation on lexical components enriches the lexical resources of the prototype like vocabulary and grammar structures. This leads the prototype to parse a reasonable amount of sentences in the corpus. While dealing with the lexicon was straight forward, the identification and extraction of appropriate semantic relations was much more involved. It was necessary, therefore, to manually develop a conceptual structure for the domain to formulate a domain-specific framework of semantic relations. The framework of semantic relations- that has resulted from this study consisted of 55 relations, out of which 42 have inverse relations. We also conducted rhetorical analysis on the corpus to prove its representativeness in conveying semantic. Finally, we conducted a topical and discourse analysis on the corpus to analyze the coverage of discourse by the prototype.**

## I. INTRODUCTION

Text to Knowledge Mapping (TKM) Prototype [1] is a domain-specific NLP system, the purpose of which is to parse instructional text and to model it with its pre-defined ontology. During development, the prototype has been tested with a limited data set from the domain instructional text on DC electrical circuit. The prototype reached a stage where its lexical components and knowledge model need to be evaluated with a representative linguistic data set, a corpus- a collection of text drawn from typical sources. Information retrieval during parsing, activation of concepts and relating them with predicate and semantic relations contribute to map and model domain-specific text on its knowledge domain. Therefore, the usability of the TKM prototype as a specialized knowledge representation tool for the domain depends on the evaluation of its lexical components like vocabulary and grammar structures, knowledge model like ontology and coverage of discourse.

An important precondition to evaluate NLP systems is the availability of a suitable set of language data called corpus as test and reference material [2]. With an extensive web-based search, we did not find any corpus for the domain DC electrical circuit. Therefore, we need to develop a representative corpus to evaluate the prototype because a representative corpus reflects the way language is used in the domain [3]. A usable corpus requires various annotations according to the scope and type of evaluation. As we intend to evaluate both the lexical components and knowledge model of the TKM prototype, the corpus should be annotated with information like Parts of Speech (POS) tagging, phrasal structure annotations, and stem word tagging. These annotations can lead us to adjust the lexical components of the prototype according to the qualitative and quantitative layers [1] [4] of its knowledge model. Thereafter, evaluation on knowledge representation of the prototype demands both development of domain-specific ontology and a generic framework of semantic relations in the domain. The evaluation helps developing a representative knowledge representation tool for the domain DC electrical circuit.

In this paper, we proposed a stochastic development procedure of a domain-specific representative corpus that is used to evaluate two major components of the TKM prototype. We presented detail procedure of corpus-based evaluation of an NLP system- that includes enriching the lexicon and morphological database, testing the parsing ability of the prototype, and the adjustment of the lexical components according to the linguistic information in the corpus. We also developed ontology according to the human conceptualization. As successful knowledge representation depends on predicate and semantic relations in the text, we developed frameworks for semantic relations with which any NLP system can read and realize text in the domain. We evaluated the coverage of discourse by the TKM prototype with a topical and discourse analysis on the corpus.

The remainder of this paper is organized as follows. In Section II, corpus-based evaluations of various NLP systems have been discussed. Section III describes the proposed procedure of representative corpus development and annotations. Section IV describes the evaluation of lexical components of the TKM prototype such as the vocabulary and grammar structure. Section V contains the outline of developing an ontology and framework for semantic relations. The section also includes the rhetorical and topical analysis. Section VI concludes the paper.

## II. RELATED WORK

A text based domain-specific NLP system can be evaluated according to the type, context or discourse of text from the domain although no established agreement has been developed on test sets and training sets [5]. Corpus is not restricted today only for researches on linguistics [6]; it is now becoming the principal resource to evaluate such domain-specific NLP systems. Many NLP systems like Saarbrucker Message Extraction System (SMES) [8] have been tested with a corpus as proper evaluation depends on a representative test set of data like corpus [7]. Corpus contains structured and variable but representative text. A corpus is said representative if the findings from it can be generalized to language or a particular aspect of language as a whole [3]. Corpus-based evaluations like MORPHIX [9] and MORPHIX++ [7] showed that the evaluation with a representative corpus results in proper adjustments. MORPHIX++ was tested with a corpus and systematic inspection revealed some necessary adjustments like missing lexical entries, discrepant morphology incomplete or erroneous single words.

NLP systems use either pre-defined or customized grammar rules. For instance, the lexical components of the TKM prototype use Combinatory Categorical Grammar (CCG) [10]. The prototype follows some specific clausal and phrasal structures according to CCG. As it follows a particular grammar, we need to adjust the grammar and phrasal structures according to the structures of text from the domain. For example, TKM prototype, on its early test, was able to parse simple sentences only [35]. This becomes a drawback if majority of text in the domain is written in compound and complex sentences. Therefore, necessary adjustment on CCG can let the prototype parse compound and complex sentences as well. In addition, NLP systems may recognize specific clausal and phrasal structure which maybe absent in domain-specific text. For example, if an NLP system uses grammars that handle one subject and one object, both parsing and knowledge extraction from domain-specific text becomes difficult if majority of the text contains more than one subject and one object. These linguistic properties of domain-specific text bring in the issue of adjustment. The lexicographical resources of such systems can be increased by analyzing linguistic patterns in domain-specific corpus. Statistical data like frequency of words, number of simple, complex or compound sentences, number of subject and object present in the sentences assist to adjust the lexical components of the systems. The grammar structure MORPHIX++ supported was not efficient in its early days. It was adjusted and extended according to the corpus used as its test suite.

The text in the corpus sometimes conveys ambiguity to a knowledge mapping prototype if its knowledge model differs from human cognition. For a sentence *a resistor is both a circuit component and a diagrammatic representation*, the role of a resistor is a component in physical connection or a component in diagram. To differentiate between them, the machine has to conceptualize the domain like human. We need semantic relations in text to conceptualize the domain. If a knowledge model is developed with domain-specific semantic relations, then machine identifies the proper role played by a concept in the domain. Semantic relations for a large domain can be obtained by developing conceptual structure of the domain with concept maps as it represents both textual and semantic relations graphically [11].

A team at Information Sciences Institute of University of Southern California was working on computer-based authoring. They suffered for an unavailability of a theory of discourse structure. Responding to this, Rhetorical Structure Theory (RST) was developed out of studies of edited or carefully prepared text from a wide variety of sources. It now has a status in linguistics that is independent of its computational uses [29]. RST is an approach to the study of text organization which conceptualizes in relational terms a domain within the semantic stratum [30]. After the formulation of RST in the 1980s, it becomes an emerging area of research for computational linguistics. It eventually draws the attention of researchers in natural language processing.

Discourse analysis helps understanding the behaviour of a domain-specific NLP system in its discourse. Corpus is a strong source of discourse analysis as linguistic and semantic relations confined in it play important role to manifest, adjust and extend systems to attune with its discourse. Researches like [31], [32], and [33] incorporated corpus in discourse analysis where emphases were given on finding linguistic relations, manual annotation and correlation of discourse structures.

## III. CORPUS DEVELOPMENT

In this section, we will discuss regarding the development approach of a domain-specific corpus, proof or its representativeness, and its annotation procedure.

### A. Development Approach

As we did not find any corpus for the domain DC electrical circuit with extensive web searches, we initiated WebBootCaT [12] to develop a representative corpus. We developed five corpora using the WebBootCaT and analyzed them by comparing the number of distinct domain-specific terms and number of distinct words present. The significant difference between these two numbers and inconsistency on the size of the corpus in Figure 1 state that web-based tools are not usable to develop domain-specific corpora.

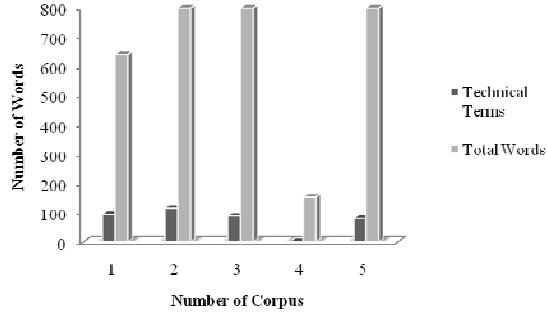

Figure 1. Inconsistency of WebBootCat to develop domain-specific corpus.

Therefore, we decided to develop the corpus manually and collected text from 141 web resources containing 1,029 sentences and 18,834 words. During the development, we left the non textual information (e.g., equations and diagrams) as the TKM prototype operates only on text.

### B. Representativeness of the Corpus

The representativeness of the corpus can be justified with a notion of saturation or closure described by [13]. At the lexical level, saturation can be tested by dividing the corpus into equal sections in terms of number of words or any other parameters. If another section of the identical size is added, the number of new items in the new section should be approximately the same as in other sections [14].

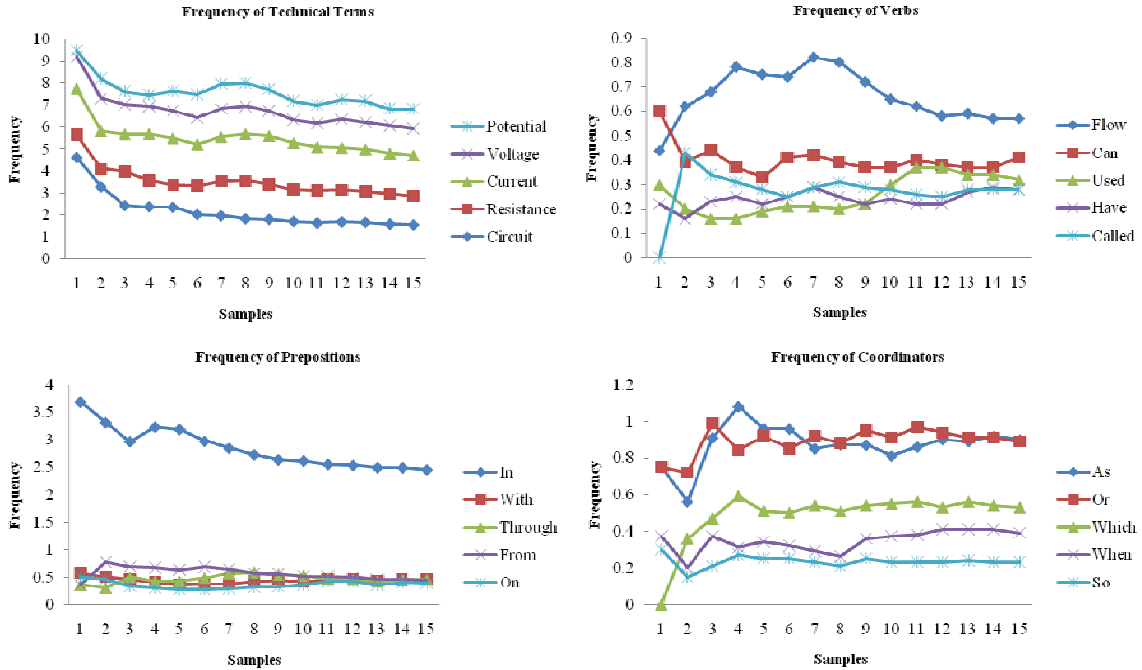

Figure 2. Representativeness of the corpus with technical terms, verbs, prepositions and coordinators

To find out the representativeness for the corpus, it has been segmented into 15 samples. Each sample is comprised of 1,267 words on average. We plotted the cumulative frequency of the most frequent technical terms in the samples.

Figure 2 depicts that the presence of the domain-specific technical terms becomes stationary after a few samples. This is one of the criteria showing the representativeness of the corpus. After a certain point, no matter how much text we add to the corpus, the frequencies of the terms are becoming stationary.

Similarly, we counted the frequency of non-technical words in the corpus and grouped them according to their parts of speech. Statistics on verbs, prepositions and co-ordinators in Figure 2 show that the corpus has been saturated after sample 11.

We also counted the frequency of types of sentences in the corpus. As the domain contains instructional text and most of which are simple sentences, it needs to be reflected on the corpus as well. Figure 3 shows that majority of the text is simple sentence (in percentage).

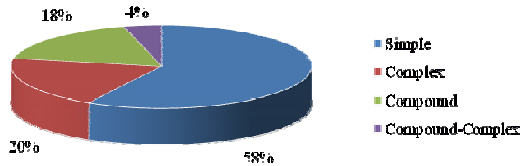

Figure 3. Sentence structure in the corpus

## C. Corpus Annotation

To annotate the corpus with POS tags, Cognitive Computation Group POS tagger [15] has been used as it works on the basis of learning techniques like Sparse Techniques on Winnows (SNOW). The corpus is annotated with nine parts of speech include noun, pronoun, verb, adverb, adjective, preposition, coordinator, determiner, and modal. The phrasal structure of the corpus has been annotated by the slash-notation grammar rules defined by CCG. We developed an XML version of the corpus with seven tags.

## IV. EVALUATION OF LEXICAL COMPONENTS

The evaluation of vocabulary and the grammar structure of the prototype are illustrated in this section. This section also refers to the efficiency in parsing and richness of lexical entries of the prototype.

### A. Evaluation of Vocabulary

The lexicon of the prototype is mapped on the unique words of the corpus. The words present both in the morphology and in corpus are called the vocabulary of the prototype. Initially, only five percent of the vocabulary was covered by the prototype (Table I).

TABLE I.
PRELIMINARY VOCABULARY COVERAGE OF THE TKM PROTOTYPE

| Words in Morphology and in Corpus | Unique Words in the Corpus | Vocabulary Coverage |
|---|---|---|
| 101 | 1,902 | 5% |

MORPHIX++, a second generation NLP system, covered 91 percent of word in the corpus developed to evaluate it. The reason behind this difference is the augmentation of the vocabulary of MORPHIX++ ran parallel with the development of the system where the main focus in case of TKM prototype was to develop an operational system first rather than increasing its vocabulary.

We used the POS tags of the corpus to populate the lexicon. We retrieved every distinct word for each distinct POS from the corpus and we simply added it if that word was absent in the lexicon. The number of added entries into the lexicon is shown in Table II. On completion of the process, the vocabulary of the prototype covers 90 percent of the corpus (Table III).

TABLE II.
AUGMENTATION OF LEXICAL ENTRIES IN THE TKM PROTOTYPE

| POS | Augmented Entries |
|---|---|
| Determiner | 19 |
| Coordinator | 5 |
| Noun and Pronoun | 2,094 |
| Adjective | 364 |
| Preposition | 71 |
| Adverb | 177 |
| Verb | 264 |

TABLE III.
VOCABULARY OF THE TKM PROTOTYPE

| Words in Morphology and in Corpus | Unique Words in the Corpus | Vocabulary Coverage |
|---|---|---|
| 1,783 | 1,902 | 90% |

### B. Evaluation of Grammar

The TKM prototype struggles to parse *modals* or *auxiliary verb* because CCG does not provide any specification to categorize modals into finite and non-finite [16]. We defined grammar formalisms for modals and adjusted the lexicon that increased the ability of the prototype to parse modals.

CCG does not have any mechanism for phrasal structures like *adjective–adjective–noun* although researches showed that numerous adjectives can be placed before a noun [17]. Except the regular adjectives, we defined grammar formalisms for noun equivalents (e.g., two *common* types of circuits), participle equivalent (e.g., the *connected* wire), gerund equivalents (e.g., the *conducting* material), and adverb equivalents (e.g., the *above* circuit is series circuit) of adjectives that increased the rate of parsing adjectives.

CCG is unable to parse sentences that start and end with a prepositional phrase [18]. For example, *in series circuit, the current is a single current*- this sentence is not parsed by CCG. In contrast, *the current is a single current in series circuit-* is sometimes parsed by CCG. The lexicon the prototype is using has nine different types of prepositions. Sometimes, it is difficult to even identify regular prepositions. For instance, *the sum of potential differences in a circuit adds up to zero voltage-* Though in regular grammars, *up* is not treated as adverbs- these are called particles where prepositions have no objects and require specific verbs with them (e.g., throw out, add up). The parsing ability of the prototype increased as we defined grammar rules for such prepositions.

*Complementizer*, although it is a form of preposition, it is not recognized by CCG. *Adverbs*, on the other hand, have a strong coverage by CCG. In many cases, adverbs sit at the end of the sentence- CCG does not provide any category to define these adverbs although it has fully featured adverb categories for other two positions of an adverb in sentence- adverbs that start a sentence or that sit in the middle of a sentence. These issues have been resolved by adding new grammar rules.

The lexicon has two categories for *coordinators*- sitting at the beginning of a sentence (e.g., since, as) and relating two clauses (e.g., and, or). CCG defined that they can be in the middle of two noun phrases only with *np\np/np* but the sentence *series and parallel circuits are the types of circuits* has the category *n\n/n* rather than *np\np/np*. CCG handles adverbs and conjunctions well but it seriously lags in handling sentences having similar verbs as in the sentence *the sum of current flowing into the junction is eventually equal to the sum of current flowing out of the junction*. The identical verbs *flowing (gerund)* appear twice with another *verb (be)* is concerning. Moreover, a verb has to be present in a sentence to form predicate argument structure but we discovered that there are sentences which do not have any verbs- *the bigger the resistance, the smaller the current.* Gerund of verb is known as noun. *Gerund* is formed by placing *ing* at the end of the verb. For example, *current flowing into a junction is equal to the current flowing out of the junction-* in this sentence, *flowing* is a gerund. Gerunds are not treated as nouns in CCG. In other words, gerunds, if treated as nouns in CCG, the sentence struggles to be parsed.

After creating grammar rules and phrasal structures and adding them into the lexicon and morphology of the prototype, the parsing ability of the prototype increased to 31 percent (Table IV). Although the prototype was tested with a limited dataset, it was unable to parse any sentence from the corpus before the evaluation.

TABLE IV.
AUGMENTATION OF LEXICAL ENTRIES IN THE TKM PROTOTYPE

| State of the Prototype | Total Sentences | Parsed Sentences | Efficiency |
|---|---|---|---|
| Preliminary | 1,029 | 0 | 0% |
| Evaluated | 981 | 300 | 31% |

We analyzed the 300 sentences parsed by the prototype and figured out the number of subject, object and verb they consist. In Figure 4, we see that the prototype works well when the number of subjects and objects in a sentence do not exceed two and when the number of verbs does not exceed one.

The inefficiency of the prototype to parse sentence is due to the absence of phrasal structures (hence, the categories). 69 percent of the sentences in the corpus have phrasal structures that are not supported by the CCG structure. It should be noted that the prototype fails to parse sentences even for absence of just one category. For example, *One simple DC circuit consists of a voltage source (battery or voltaic cell) connected to a resistor –* this sentence is not parsed by the prototype for the absence of category of conjunction *or* (np\n/np) and for the category of verb *connected* (s\np/pp). In the corpus, these absent categories are identified so that modification of the lexicon becomes easier.

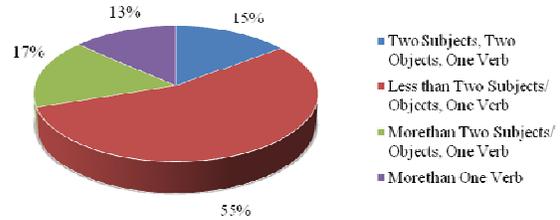

Figure 4. Number of subjects, objects, and verbs in the sentences parsed by TKM prototype.

The inefficiency of the prototype to parse sentence is due to the absence of phrasal structures (hence, the categories). 69 percent of the sentences in the corpus have phrasal structures that are not supported by the CCG structure. It should be noted that the prototype fails to parse sentences even for absence of just one category. For example, *One simple DC circuit consists of a voltage source (battery or voltaic cell) connected to a resistor –* this sentence is not parsed by the prototype for the absence of category of conjunction *or* (np\n/np) and for the category of verb *connected* (s\np/pp). In the corpus, these absent categories are identified so that modification of the lexicon becomes easier.

V. EVALUATION OF KNOWLEDGE MODEL

In this section, we will discuss the procedure of developing a domain-specific ontology and framework for semantic relations. The results of rhetorical, topical, and discourse analysis are also outlined in this section.

*A. Ontology and Framework for Semantic Relations*

From the 300 parsed sentences, the prototype is able to map only 10 percent of the sentences effectively on its pre-built ontology. We investigated the ontology and found that it was not developed according to a representative data set like our corpus. We decided to develop ontology for the domain-specific corpus that helps to adjust the knowledge model of the TKM prototype.

We developed the ontology in a similar way human conceptualizes a domain. In conjunction with the development of the ontology for the domain, we developed a framework for semantic relations. The framework is built upon the framework proposed by FACTOTUM thesaurus [19] [20]. These semantic relations help to represent hierarchical knowledge apart from predicate information.

We conceptualized every sentence in the corpus manually. The outcome of the conceptualization led us to develop concepts and relations among them and graphically represented them as concept maps with Cmap Tools [21].

To illustrate this procedure, for the sentence *One simple DC circuit consists of a voltage source (battery or voltaic cell) connected to a resistor*, we firstly conceptualized the sentence in the following manner-
1. DC circuit has voltage source as its component.
2. Battery and voltaic cell are voltage sources.
3. Battery and voltaic cell have similarity.
4. Voltage source can be connected to resistor.
5. DC circuit has resistor as its component.
6. As they all are satisfying the properties of a circuit, DC circuit is a type of circuit.

We used this information to develop base level concept maps that represent the predicate relations in the text. To develop higher level concept maps, we require to group concepts and to find relations among the groups. For this particular sentence, we defined groups named *circuit* and *circuit component*. We assigned *DC Circuit* and *Circuit* to the group *Circuit* and the rest of the concepts to the group *circuit component*. We can also find a relation between these two groups- circuit *is made of* circuit components. For a sentence *Resistors in the diagram are in parallel*- the concept *resistor* would be assigned to group of concepts called *Diagrammatic Notation* rather than *Circuit Components*. This process of grouping the concepts from the base level concept maps and finding relations among them produced four levels of concept maps for the corpus. The conceptual structure of the domain is comprised of all these concept maps resulted from human conceptualization at four different levels.

The predicate relations in the sentence are as follows-
1. DC Circuit *Have Component* Voltage source
2. Battery *Type Of* voltage source
3. Voltaic cell *Type Of* voltage source
4. Battery *Is* Voltaic Cell
5. Voltage Source *Connected To* Resistor
6. Battery *Connected To* Resistor
7. Voltaic Cell *Connected To* Resistor
8. DC Circuit *Have Component* Resistor
9. DC Circuit *Type of* Circuit

These relations are then analyzed to initiate developing the framework for the semantic relations in the text. The analysis provides us the following semantic relations-
1. Relation which describes parts that are physically related (e.g., Have Component)
2. Relation which describes hyponymy (e.g., Type Of), and synonymy (e.g., Is) that are similar
3. Relation which describes hierarchy or class (e.g., Type Of)
4. Relation which describes spatial relations (specifically location of objects) (e.g., Connected To)

As we represent knowledge by conceptualization followed by mapping linguistic information on knowledge model, it will allow the prototype to map knowledge from the text onto the ontology efficiently. For example, the prototype now can provide the user knowledge like *voltage source* is a physical part of the *DC circuit*- which is not stated in the sentence literally but semantically.

As we developed conceptual structure for the corpus with the Cmap Tool, the total number of concepts and relations increases but number of new concepts and relations decreases. On completion, the number of concepts and relations are plotted against the corpus size. Figure 5 shows the cumulative increment of the number of concepts and relations. We see a plateau showing that the number of concepts and relations are becoming stationary.

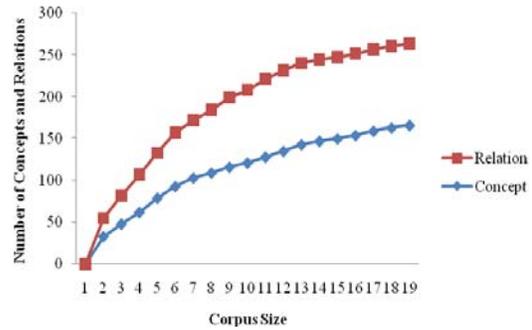

Figure 5.  Graph to show that the number of concepts and relations in the corpus is becoming stationary.

We also plotted number of new concepts and relations against the corpus size (Figure 6). The plateau in Figure 6 shows that the number of new concepts and relations are becoming stationary. These two observations led us to a decision that if we put semantic relations in the corpus into a framework, then it will be representative.

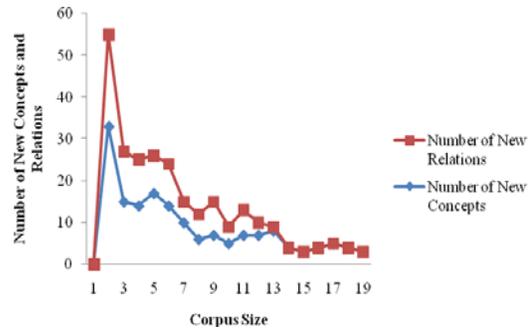

Figure 6.  Graph to show that the number of new concepts and relations in the corpus is becoming stationary.

We found 97 predicate relations and 166 concepts in the corpus and we developed Tier 2 of our framework (Table V) to support these relations. Afterwards, we grouped level 0 concepts and relations to produce level 1 of concept maps. As we came across new predicate relations, we created Tier 1 of our framework to support the semantic relations in Tier 2. These two tiers of semantic relations comprise the domain-specific framework for semantic relations and can be supportive to all the predicate relations of the domain. In essence, the level 0 concept maps have the predicate relations and the semantics conveyed by them are supported by relations in Tier 2. Predicate relations in level 1 and level 2 concept maps are supported by Tier 1 semantic relations. The ontology along with the concept maps is depicted in [34].

TABLE V.
FRAMEWORK FOR SEMANTIC RELATIONS IN THE CORPUS

| Relation Category | Tier 1 Semantic Relations | Tier 2 Semantic Relations | Predicate Relations | Inverse Predicate Relations |
|---|---|---|---|---|
| Predicate Relations | Hierarchy | | Have type | Type of |
| | Physically Related | Parts | Have component | Component of |
| | | Constituent Material | Make, Produce | Made of, Produced by |
| | Spatial Relations | Location of Objects | Take place between, Connected to, Flows through, Have direction | Direction of |
| | | Location of Activities | Transfer, Find, Divide, Commence from | Transferred by, Found by, Divided by, End to |
| | Causally/ Functionally Related | Effect/ Partial Cause | Affect, Cause, Vary in, Resist, Force, Limit, Opposite to, Related to | Affected by, Caused by, Resisted by, Forced by, Limited by |
| | | Production/ Generation | Produce | Produced by |
| | | Destruction | Collide, Melt | Collided by, Melted by |
| | | Manifestation | Represent | Represented by |
| | | Conversion | Convert, Convertible to | Converted by, Convertible from |
| | Instrumental Function/ Usage | Functions | Carry, Measure, Supply, Share, Depend on, Protect, Absorb | Carried by, Measured by, Supplied by, Shared by, Depended by, Protected by, Absorbed by |
| | | Use | Use, Do not use | Used by, Not used by |
| | Human Role | | Deal with | Dealt by |
| | Conceptually Related | Topic | Govern | Governed by |
| | | Representation | Represent, Characterize | Represented by, Characterized by |
| | | Property | Have state, Have unit, Have source, Have Magnitude, Have Terminal | State of, Unit of, Source of, Magnitude of, Terminal of |
| | Similarity | Synonymy | Is, Referred to | Is |
| | | Hyponymy | Have type | Type of |
| | Quantitative Relations | Numerical Relations | Proportional, Inverse proportional to, Gain, Lose, Do not gain, Do not lose | |
| Instantiation | | | Have instance | Instance of |
| Extension | | | Have Extension | Extension of |

*B. Rhetorical Analysis*

To find the stereotypical relations in the domain, RST proposed by Mann & Thompson [22] is used as a descriptive tool. Research work like Rosner & Stede [23] and Vander Linden [24] also used this framework for the rhetorical analysis in their corpus. We used a framework based on the work of Hunter [25] who outlines the structural model of content of information for second language learning materials proposed within the frame of machine-mediated communication [26]. This framework defined text structures, textual expressions and information structures within domain-specific text.

One common characteristic of expository text is that they use text structures. *Text structures* refer to the semantic and syntactic organizational arrangements used to present written information. Text structure used in the analysis includes *introduction, background, methodologies, results, observations,* and *conclusions.*

*Textual Expressions* are relations that describe the nature of a sentence at phrase level. It eventually outlines the type of the sentence. These all are mononuclear relations- the relations do not depend on the semantic of the adjacent sentences. We used the following textual expressions for the analysis- *common knowledge, cite, report, explanation, claim, evaluation inference,* and *decision.*

*Information structures* are used at both phrase level and sentence level in the analysis. We analyzed the meaning of the sentence and its impact on other juxtaposed sentences with relations like *description, classification, comparison, sequence, cause-effect,* and *contrast.*

We used RSTTool [27] to annotate the corpus with the rhetorical relations. The procedure shows that the corpus has 2,701 relations grouped into 19 rhetorical relations (Table VI).

Higher means of relations *background* (13 percent) and *observations* (7 percent) are significant as the corpus contains instructional text and instructional text mostly describes background and observation of events [28]. The analysis also shows that most of the text are descriptive (30 percent) and presented as report (20 percent) thus proved the representativeness of the corpus in case of containing semantic relations. Qualitative layer of the prototype deals with the causal relationship between concepts and a significant amount of *cause-effect* relation (2.14 percent) is of particular interest for us to deal with. We found that the prototype is able to map about 70 percent of the causal relations in the text.

TABLE VI.
RHETORICAL RELATIONS IN THE CORPUS

| Rhetorical Structures | Rhetorical Relations | Appearance | Mean |
|---|---|---|---|
| Text Structures | Introduction | 117 | 4.33% |
| | Background | 356 | 13.19% |
| | Methodologies | 60 | 2.22% |
| | Results | 51 | 1.89% |
| | Observations | 180 | 6.66% |
| | Conclusions | 42 | 1.55% |
| Textual Expressions | Common Knowledge | 74 | 2.74% |
| | Report | 545 | 20.18% |
| | Explanation | 192 | 7.11% |
| | Claim | 85 | 3.15% |
| | Evaluation | 3 | 0.11% |
| | Inference | 13 | 0.48% |
| | Decision | 59 | 2.18% |
| Information Structures | Description | 817 | 30.25% |
| | Classification | 13 | 0.48% |
| | Comparison | 25 | 0.93% |
| | Sequence | 34 | 1.26% |
| | Cause-effect | 58 | 2.14% |
| | Contrast | 17 | 0.63% |
| | Total | 2,701 | 100% |

*C. Topical Analysis*

We intend to analyze the topical progression of the corpus as the prototype both handled and failed to handle text on various topics. The analysis will help us to determine the context and discourse awareness of the prototype. The prototype is not developed to parse and map text of any particular topic or context and it should represent the whole domain. Since we have the representative corpus, the topical analysis of the prototype can help us understand the topical coverage of its context and discourse.

First, we annotated the corpus with three types of topical progressions- parallel progression, sequential progression, and extend parallel progression. As the topic in the text progresses onwards, we indented the text of the corpus according to the type of progression it belongs to. For example, indentation <1a> is the starting topic, indentation <2> is the sequential topic originated from <1a>, indentation <3> is the sequential topic originated from <2>, and indentation <1b> is the extended parallel topic of <1a> (Figure 7). On completion, we found six indentations of topical progression in the corpus.

<1a> Putting more resistors in the parallel circuit decreases the total resistance because the electricity has additional branches to flow along and so the total current flowing increases.
    <2> This is very useful because it means that we can switch the lamp on and off without affecting the other lamps.
        <3> The brightness of the lamp does not change as other lamps in parallel are switched on or off.
<1b> For this reason, lamps are always connected in parallel.

Figure 7. Annotation of the corpus with topical progression.

Second, we counted total number of sentences in each indentation of the corpus. As we expected, indentation 1 covers the most of the corpus and indentation 6 has the least number of sentences. We also counted number of sentences TKM prototype handled in each indentation to find its topical coverage. The more the topic progresses away from the context, the possibility of not understanding the context increases but the prototype showed that even if the topic is six indentations away from the original context, it can represent the knowledge (Table VII). The prototype efficiently handled language and knowledge on topics that are four, five, and six indentations away from the original context with 38, 36, and 33 percent coverage, respectively. However, topics nearer to the starting context are covered relatively low with 25 and 26 percent.

TABLE VII.
TOPICAL ANALYSIS OF THE CORPUS AND TKM PROTOTYPE

| Indentation | Number of Sentences | Corpus Coverage | Number of Sentences handled by the Prototype | Topical Coverage by the Prototype |
|---|---|---|---|---|
| 1 | 641 | 66% | 197 | 31% |
| 2 | 259 | 22% | 65 | 25% |
| 3 | 86 | 7% | 22 | 26% |
| 4 | 26 | 3% | 10 | 38% |
| 5 | 11 | 1% | 4 | 36% |
| 6 | 4 | 1% | 2 | 33% |

*D. Discourse Analysis*

The discourse of the prototype contains high level concepts developed during the progress of ontology. High level concepts are those that are related with Tier 1 and Tier 2 semantic relations of our framework and convey knowledge rather than predicate information. According to our research, the domain has 12 high level concepts shown in Table VIII. The table is organized in descending order according to the number of concepts in discourse. We semi-automatically analyzed the corpus and found that the high level concepts of the ontology are present 4,120 times in the corpus- this is the discourse of the prototype. Moreover, we also found that the high level concepts of the domain are present 969 times in the sentences that the prototype can handle- which is the discourse covered by the prototype. If we divide the discourse coverage of prototype by the total number of concepts in the discourse, then we will find the discourse covered by the prototype. In this case, Table VIII shows that the discourse coverage of the TKM Prototype is 24 percent.

If we consider individual high level concepts, then *Units* and *Measuring Instruments* are the areas of discourse the prototype covers, mostly, with 48 percent of coverage. *Rules* is next to them with 28 percent of coverage. The prototype covers only 15 percent of the discourse of *Electrical Process* though the discourse is significant in the domain. *Environmental Factors*, a narrower high level concept, is next to it in case of less discourse coverage by the prototype.

TABLE VIII.
DISCOURSE ANALYSIS OF THE TKM PROTOTYPE

| High Level Concepts | Coverage of Prototype | Concepts in Discourse | Difference with Discourse | Discourse Coverage | Deviation |
|---|---|---|---|---|---|
| (1) Electrical Quantity | 335 | 1433 | 1098 | 24% | 77% |
| (2) Circuit Components | 154 | 685 | 531 | 23% | 78% |
| (3) Diagrammatic Notation | 94 | 482 | 388 | 20% | 81% |
| (4) Electrical Process | 63 | 442 | 379 | 15% | 86% |
| (5) Electrical Device | 83 | 313 | 230 | 27% | 74% |
| (6) Units | 100 | 211 | 111 | 48% | 53% |
| (7) Atomic Level | 31 | 161 | 130 | 20% | 81% |
| (8) Circuits | 30 | 140 | 110 | 22% | 79% |
| (9) Environmental Factors | 20 | 110 | 90 | 19% | 82% |
| (10) Measuring Instrument | 46 | 96 | 50 | 48% | 53% |
| (11) Rules | 13 | 47 | 34 | 28% | 73% |
| (12) Materials | 4 | 21 | 17 | 20% | 81% |
| Total | 969 | 4,120 | 3,151 | 24% | 77% |

We also analyzed the deviation of the prototype from the discourse. First, we measured the difference of the coverage of the prototype and coverage in discourse. Then, we measured the deviation- the difference with discourse divided by the concepts in discourse. This deviation is the measure of unawareness in discourse- how much of the discourse the prototype failed to pursuit. The data show that the prototype is strong to represent knowledge from the discourse of *Units* and *Measuring*

*Instrument* (both 53 percent). The prototype has the overall deviation from the discourse of 77 percent- means its discourse awareness is 23 percent.

We plotted the presence of high level concepts in the discourse and the coverage of the discourse by the prototype in Figure 8. The difference between the coverage of discourse by the prototype and the discourse itself is depicted with vertical lines. From Figure 8 and Table VIII, we see that the difference is proportional to each other from *Electrical Quantity* to *Electrical Processes* and then a sudden rise in case of *Electrical Device* and *Units* indicates that most of the simple sentences in the corpus are situated in this area. Another smooth maintenance of difference between the discourse and the coverage of discourse is manifested from concepts *Atomic Level* to *Environmental Factors*. The prototype shows efficiency in knowledge representation in *Measuring Instruments* that indicates to the possibility of having understandable knowledge for the prototype lies in this discourse with suitable linguistic and semantic arrangement.

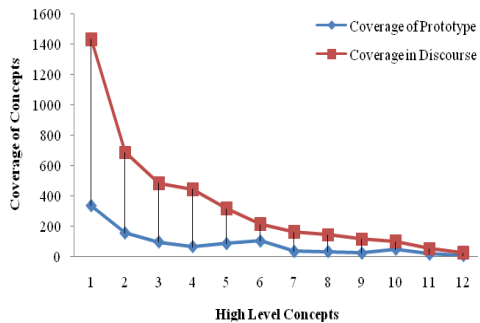

Figure 8. Difference between the discourse and the coverage of discourse by the prototype.

## VI. CONCLUSIONS

In this paper, we presented a corpus-based evaluation of lexical components and knowledge model of a domain-specific Text to Knowledge Mapping prototype. We developed a domain-specific corpus and proved its representativeness in linguistic elements with stochastic approach and its soundness in semantic features with rhetorical analysis. The representative corpus, with enriched multimodality, can be used as a reference in text summarization, for context and discourse analysis, and for developing ontology. The linguistic resources of the corpus have been used to evaluate and adjust lexical components of the prototype like vocabulary and grammar. This evaluation led the prototype to parse reasonable amount of domain-specific text. During evaluation on knowledge model, we developed a domain-specific ontology and a framework for semantic relations associated with it. We conducted topical and discourse analysis on the prototype to see its context awareness and the performance of the prototype is satisfactory. However, limited conceptual acquisition of the prototype refers to limited knowledge representation and demands a framework for domain-specific linguistic relations.

Using the domain-specific corpus, a generic corpus parsing and lexical component analysis tool is developed [36] that extracts lexical information from any XML corpus and store the information in database. The corpus also contributed in domain-specific text summarization and the result of summarization was satisfactory [37].

**Rushdi Shams** was born in Khulna, Bangladesh on January 3, 1985. He pursued his M.Sc. in Information Technology from the University of Bolton, United Kingdom in 2007 and his B.Sc. (Engineering) in Computer Science and Engineering from Khulna University of Engineering & Technology (KUET), Bangladesh in 2006. He has major fields of study like intelligent systems, internet security, data warehouse, IT management and artificial intelligence.

He is currently a Lecturer in the Department of Computer Science and Engineering, KUET. Formerly, he was a Research Intern at M3C Laboratory, University of Bolton. So far, he has publications in the area of Ad hoc Networks and Knowledge Processing. He supervised undergraduate theses on diversified fields like knowledge processing, wireless sensor networks, corpus linguistics, and web engineering. Currently, he is developing frameworks for web 3.0 and acquisition and machine representation of commonsense knowledge. His current research interests are Knowledge and Language Processing, Computational Linguistics, and Wireless Networks.

Mr. Shams is an Associate Member of Institute of Engineers, Bangladesh (IEB).

**Adel Elsayed** pursued his Ph.D on Applied Optimal Control from Loughborough University, UK in 1985. His M.Sc is in Electronic Control Engineering from Salford University, UK in 1975 preceded by a B.Sc. in Communications Engineering from Libya University, Libya in 1972. He has a diversified field of study from signal processing to multimodal communication and from communications to intelligent systems and human-computer interaction.

He joined the University of Bolton towards the end of 2001. His main objective was to establish a research base by building on his work on multimodal communication, a new line of research that he has started few years earlier. His research at Bolton has started as he set up the "Active Presentation Technology" Lab (APT Lab). Since then, he diversified into investigating the underpinning knowledge structures that support human-machine communication. This led to researching information structures and their applications. Consequently, the scope of his work grew out of the narrow area of active presentation technology into the wider scope of Machine-Mediated Communication, hence the new name of the research lab. He has esteemed numbers of conference and journal publications. Besides these, his research interests are cognitive tools, knowledge and language processing, and speech technology.

Dr. Elsayed established M3C as international workshop attached to well known IEEE International Conference on Advanced Learning Technologies (ICALT). He acts as guest editor for special issues of many well known journals as well as reviewer in number of international conferences.

**Quazi Mah- Zereen Akter** was born on November 25, 1985 in Dhaka, Bangladesh. She is now completing her M.Sc. in Computer Science and Engineering from the University of Dhaka, Bangladesh. She completed her B.Sc. (Engineering) in Computer Science and Engineering from the University of Dhaka in 2008. Bioinformatics, artificial intelligence, computer vision, distributed database systems, digital system design, and VLSI are her major fields of study.

She is currently working on her Masters thesis with reversible logic and computational intelligence. Besides Intelligent Systems, her research interests are Bioinformatics, Management Information Systems, Reversible Logic, and Logic Simplification and Minimization.